\documentclass[letterpaper, 10 pt, conference]{ieeeconf}
\usepackage{times}
\usepackage{color, colortbl, xcolor}
\usepackage{multicol}
\usepackage[bookmarks=true]{hyperref}
\usepackage{amsfonts}
\usepackage{algorithmic}
\usepackage[ruled,vlined,titlenotnumbered]{algorithm2e} 
\usepackage{amsmath}
\usepackage{dsfont}
\usepackage{subfigure}
\usepackage[pdftex]{graphicx}   
\usepackage{graphicx}
\usepackage{balance}

\graphicspath{{./figs/}}    
\DeclareGraphicsExtensions{.pdf,.png}  

\usepackage[textfont=md,font=footnotesize]{caption}

\usepackage[%
    style=numeric-comp,
    sorting=none,
    backend=biber,
    sortcites=true,
    doi=false,
    url=false,
    firstinits=true,
    hyperref,
    isbn=false,
    eprint=false,
    maxcitenames=3, 
    minbibnames=3, 
    maxbibnames=4, 
    block=none]
    {biblatex}
    \renewbibmacro{in:}{}
    \AtEveryBibitem{%
  	\clearlist{language}%
  	\clearfield{pages}%
	}
\bibliography{library}

\newbibmacro{string+doiurlisbn}[1]{%
  \iffieldundef{doi}{%
    \iffieldundef{url}{%
          #1
    }{%
      \href{\thefield{url}}{#1}
    }%
  }{%
    \href{https://dx.doi.org/\thefield{doi}}{#1}
  }%
}

\DeclareFieldFormat{title}{\usebibmacro{string+doiurlisbn}{\mkbibemph{#1}}}
\DeclareFieldFormat[article,incollection]{title}%
    {\usebibmacro{string+doiurlisbn}{\mkbibquote{#1}}}

\usepackage{ifthen}
\newboolean{include-notes}
\setboolean{include-notes}{true}

\newcommand{\jfnote}[1]%
    {\ifthenelse{\boolean{include-notes}}
    {\textcolor{blue}{\textbf{Jaime: #1}}}{}}
\newcommand{\ebnote}[1]%
    {\ifthenelse{\boolean{include-notes}}
    {\textcolor{purple}{\textbf{Eli: #1}}}{}}
\newcommand{\adnote}[1]%
    {\ifthenelse{\boolean{include-notes}}
    {\textcolor{orange}{\textbf{Anca: #1}}}{}}
\newcommand{\dsnote}[1]%
    {\ifthenelse{\boolean{include-notes}}
    {\textcolor{cyan}{\textbf{Dorsa: #1}}}{}}
\newcommand{\remove}[1]%
    {\textcolor{red}{#1}}  

\usepackage{lipsum}

\newcommand\blfootnote[1]{%
  \begingroup
  \renewcommand\thefootnote{}\footnote{#1}%
  \addtocounter{footnote}{-1}%
  \endgroup
}

\newcommand{\A}{\mathcal{A}}

\renewcommand{\H}{\mathcal{H}}

\newcommand{\U}{\mathcal{U}}

\newcommand{\sdyn}{\phi}

\title{\LARGE \bf
Hierarchical Game-Theoretic Planning for Autonomous Vehicles
}

\author{
    Jaime F. Fisac$^{*1}$ \and
    Eli Bronstein$^{*1}$ \and
    Elis Stefansson$^{2}$ \and
    Dorsa Sadigh$^{3}$ \and
    S. Shankar Sastry$^{1}$ \and
    Anca D. Dragan$^{1}$
}

\begin{document}
\maketitle
\thispagestyle{empty}
\pagestyle{empty}
\blfootnote{
\hspace{-0.52cm}
$^{1}$Department of Electrical Engineering and Computer Sciences.
University of California, Berkeley, United States. \newline
$^{2}$Department of Mathematics, KTH Royal Institute of Technology, Sweden. \newline
$^{3}$Computer Science Department, Stanford University, United States.
    Email:
    \{%
    \href{mailto:ebronstein@berkeley.edu}{ebronstein},
    \href{mailto:shankar_sastry@berkeley.edu}{shankar\_sastry},
    \href{mailto:jfisac@berkeley.edu}{jfisac},
        \href{mailto:anca@berkeley.edu}{anca}%
    \}@berkeley.edu,
    \href{mailto:elisst@kth.se}{elisst@kth.se}
    \href{mailto:dorsa@cs.stanford.edu}{dorsa@cs.stanford.edu}
\newline
$^*$The first two authors contributed equally to this work.
\newline
This work was partially supported by an NSF CAREER award and a Ford URP. We thank NVDIA for hardware support.
\newline
The authors would like to thank Karl H. Johansson for helpful discussions and valuable input on vehicle coordination.
}%
%
%
\begin{abstract}

The actions of an autonomous vehicle on the road affect and are affected by those of other drivers,
whether overtaking, negotiating a merge, or avoiding an accident. 
This mutual dependence, best captured by dynamic game theory, creates a strong coupling between the vehicle's planning and its predictions of other drivers' behavior,
and constitutes an open problem with direct implications on the safety and viability of autonomous driving technology.
Unfortunately, dynamic games are too computationally demanding to meet the real-time constraints of autonomous driving in its continuous state and action space.
In this paper, we introduce a novel game-theoretic trajectory planning algorithm for autonomous driving,
that enables real-time performance by hierarchically decomposing
the underlying dynamic game
into a long-horizon ``strategic" game with simplified dynamics and full information structure,
and a short-horizon ``tactical" game with full dynamics and a simplified information structure.
The value of the strategic game is used to guide the tactical planning,
implicitly extending the planning horizon, pushing the local trajectory optimization closer to global solutions, and, most importantly, quantitatively accounting for the autonomous vehicle and the human driver's ability and incentives to influence each other.
In addition, our approach admits non-deterministic models of human decision-making,
rather than relying on perfectly rational predictions.
Our results showcase richer, safer, and more effective autonomous behavior in comparison to existing techniques.

\end{abstract}

\section{Introduction}\label{sec:introduction}

Imagine you are driving your car on the highway and, just as you are about to pass a large truck on the other lane, you spot another car quickly approaching in the wing mirror.
Your driver's gut immediately gets the picture:
the other driver is trying to squeeze past and cut in front of you at the very last second, barely missing the truck.
Your mind races forward to produce an alarming conclusion:
it is too tight---yet the other driver seems determined to attempt the risky maneuver anyway.
If you brake immediately, you could give the other car enough room to complete the maneuver without risking an accident; if you accelerate, you might close the gap fast enough to dissuade the other driver at the last second.

Driving is fundamentally a game-theoretic problem, and safety depends on getting the solution right.
However, most approaches in the literature follow a ``pipeline" approach that generates predictions of the trajectories of human-driven vehicles and then feeds them to the planning module as unalterable moving obstacles \cite{Carvalho2013,Vitus2013,Luders2010,Hermes2009}.
This can lead to both excessively conservative and in some cases unsafe behavior~\cite{felton2018google},
a well-studied issue in the robotic navigation literature known as the ``frozen robot" phenomenon~\cite{Trautman2010a}.

Recent work has addressed this by modeling human drivers as utility-driven agents who will plan their trajectory in response to the autonomous vehicle's internal plan.
The autonomous vehicle can then select a plan that will elicit the best human trajectory in response~\cite{Sadigh2016,Sadigh2016b}.
Unfortunately, this treats the human as a pure \emph{follower} in the game-theoretic sense, effectively inverting the roles in previous approaches.
That is, the human is assumed to take the autonomous vehicle's future trajectory as immutable and plan her own fully accommodating to it, rather than try to influence it.
Further,
the human driver must be able to observe, or exactly predict, the future trajectory planned by the autonomous vehicle, which is hardly realistic for anything other than very short planning horizons.
As we will discuss in our results,
this can lead to undesirably aggressive vehicle behavior.

\begin{figure}[t!]
   \centering
    \includegraphics
        [scale = 0.29]
        {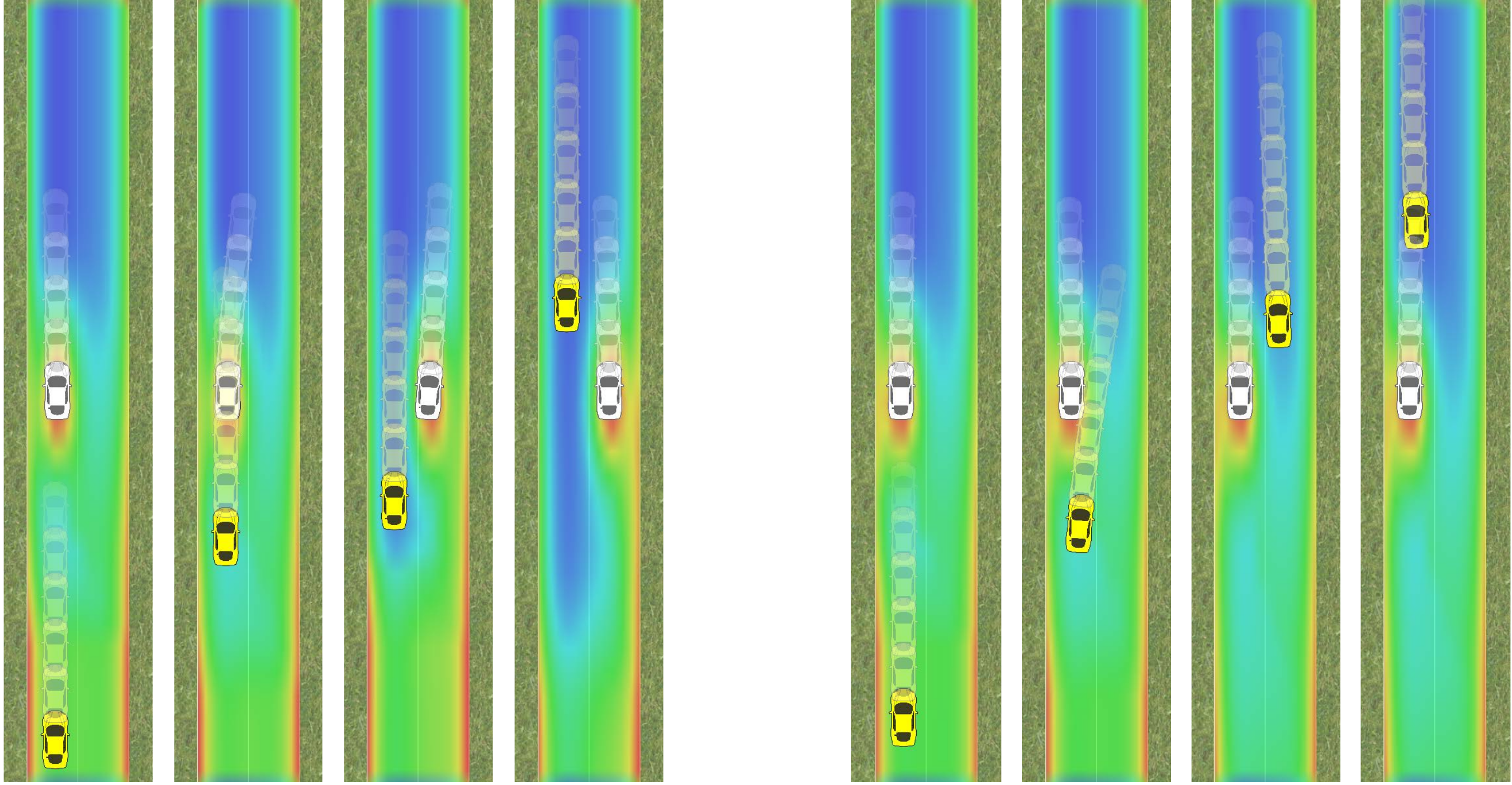}
    \caption{
        Demonstration of our hierarchical game-theoretic planning framework on a simulated overtaking scenario.
        The autonomous vehicle (yellow) is initially driving faster than the human-driven car (white).
        The heatmap displays the hierarchical planner's strategic value, ranging from red (low value) to blue (high value),
        which accounts for the outcome of possible interactions between the two vehicles.
        (a) As the autonomous vehicle approaches, it is incentivized to pressure the human to change lanes and let it pass (note the growth of a high-value region directly behind the human in the left lane).
        (b) If the human does not maneuver, the autonomous vehicle chooses to change lanes and overtake, following the higher values in the right lane.
        }
    \label{fig:4d_overtaking}
\end{figure}

In this work, we introduce 
a hierarchical game-theoretic framework
to address the mutual influence between the human and the autonomous vehicle while maintaining
computational tractability.
Our framework hinges on the use of a coupled interaction model in order to plan for horizons of multiple seconds, during which drivers can affect each other's behavior through their actions over time.
We do this by computing the optimal value and strategies for a dynamic nonzero-sum game with a long horizon (typically a few seconds) and a full closed-loop feedback information structure~\cite{Basar1984,Simaan1973}.
In order to maintain tractability, we propose solving this long-horizon game using simplified dynamics, which will approximately capture the vehicles' ability to execute different trajectories.
The resulting long-term value, which captures the expected outcome of the strategic interaction from every state,
can then be used as an informative terminal component in the objective function used in a receding-horizon planning and control scheme.
This low-level planner can use a higher-fidelity representation of the dynamics, while only planning for a shorter time horizon (typically less than one second) during which simplifications in the interaction have a less critical effect 
~\cite{Sheridan1966a,Peng1993,MacAdam2003}.%

Our framework therefore hierarchically combines:
\begin{itemize}
    \item A \emph{strategic} (high-level) planner that determines the outcome of long-term interactions using simplified dynamics and fully coupled interaction.
    \item A \emph{tactical} (low-level) planner that computes short-term vehicle trajectories using high-fidelity dynamics and simplified interaction, informed by the long-term value computed by the strategic planner.
\end{itemize}

By more accurately capturing the information structure in the interaction between the autonomous vehicle and other drivers and using a more tractable dynamical model,
the hierarchical framework makes it possible to reason farther into the future than most receding-horizon trajectory planners.
The high-level game value informs the trajectory optimization as a terminal cost, implicitly giving it an approximate insight into the longer time scale (in a similar spirit to a variety of planning schemes, e.g. \cite{Silver2016a}).
In addition, since this strategic value is computed globally via dynamic programming, it can help mitigate the local nature of most trajectory optimization schemes, biasing them towards better~solutions.

An important strength of our framework is that the strategic planner does not require using a deterministic model of the human, such as an ideal rational agent,
but instead allows a variety of models including probabilistic models such as noisy rationality, commonly used in inverse optimal control (also inverse reinforcement learning)~\cite{Ziebart2008,Finn2016}.
In addition, the framework is agnostic to the concrete planner used at the tactical level:
while we demonstrate our approach with an optimizer based on ~\cite{Sadigh2016}, this could be replaced with other methods, including deep closed-loop prediction models, such as~\cite{Schmerling2018}, by introducing the strategic value as a terminal cost term in their objective function.
Therefore, the method proposed here should not be seen as competing with such planning schemes, but rather as enhancing them.

Importantly, solving the underlying dynamic game does not imply that the autonomous vehicle will be more aggressive---its driving behavior will ultimately depend on the optimization objective specified by the system designer, which may include terms encoding comfort and safety of other road users.
With adequate objective design, our framework can enable safer and more efficient autonomous driving by planning with a more accurate model of interactions.

\section{Dynamic Game Formulation} \label{sec:formulation}

We consider a single%
\footnote{%
    While extension of our formulation and solution to $N$ players is well-defined (and relatively straightforward) in \emph{theory},
    in practice computing the solutions requires exponential
    computation in the number of vehicles involved.
    We thus limit the scope of this work to pairwise interactions, and note that recent prediction approaches~\cite{fisac2018probabilistically} may enable viable extensions.
}
human driver $H$ and a single autonomous system $A$ in control of their respective vehicles.
The dynamics of the joint state $x^t\in\mathcal{X}\subset\mathbb{R}^{n}$ of the vehicles in the world,
which we assume to be fully observable, are
\begin{equation}\label{eq:dynamics}
    x^{t+1} = f(x^t,u_A^t,u_H^t)
    \enspace,
\end{equation}
where
$u_i^t\in\mathcal{U}_i\subset\mathbb{R}^{m_i}$ is the driving control action for each $i\in\{A,H\}$ at time step $t$; we assume $\U_i$ is compact.

The autonomous system is attempting to maximize an objective that depends on the evolution of the two vehicles over some finite time horizon, namely a cumulative return:
\begin{equation}\label{eq:return}
    R_A(x^{0:N}, u_A^{0:N}, u_H^{0:N}) = \sum_{t=0}^{N} r_A(x^t, u_A^t, u_H^t)
    \enspace.
\end{equation}
The reward function $r_A$ captures the designer's specifications of the vehicle's behavior and may encode aspects like fuel consumption, passenger comfort, courteousness, time efficiency, and safety .
Some of these aspects (crucially safety) may depend on the joint state of the two vehicles;
the reward function may also explicitly depend on the human driver's actions (the designer may, for instance, decide to penalize it for causing other vehicles to maneuver abruptly).
The autonomous vehicle therefore needs to reason about not only its own future actions, but also those of the human driver.

We assume that the autonomous vehicle has some predictive model of the human's actions as a function of the currently available information (the joint state, and possibly the autonomous vehicle's current action).
The coupling in the planning problem is then explicit.
If the system models the human as exactly or approximately attempting to maximize her own objective function,
the coupling takes the form of a dynamic game,
in which each player acts strategically as per her own objective function accounting for the other's possible actions.
Since both players observe the current state at each time, this dynamic game has closed-loop feedback information structure, and optimal values and strategies can be computed using dynamic programming~\cite{Basar1984,Simaan1973a}.

Unfortunately, deriving these strategies can be computationally prohibitive due to the exponential scaling of computation with the dimensionality of the joint state space (which will be high for the dynamical models used in vehicle trajectory planning).
However, we argue that successfully reasoning about traffic interactions over a horizon of a few seconds does not require a full-fidelity model of vehicle dynamics,
and that highly informative insights can be tractably obtained through approximate models.
We further argue that it is both useful and reasonable to model human drivers as similarly reasoning about vehicle interactions over the next few seconds
without needing to account for fully detailed dynamics.
This insight is at the core of our solution approach.

\section{Hierarchical Game-Theoretic Planning}\label{sec:solution}

We propose a hierarchical decomposition of the interaction between the autonomous vehicle and the human driver.
At the high level, we solve a dynamic game representing the long-horizon interaction between the two vehicles through approximate dynamics.
At the low level, we use the computed value function as an approximation of the best long-horizon outcome achievable by the autonomous vehicle from each
state,
and incorporate it in the form of a guiding terminal term in the short-horizon trajectory optimization, which is solved in a receding-horizon fashion with a high-fidelity model of the vehicles' dynamics.

\subsection{Strategic planner: Closed-loop dynamic game}\label{subsec:strategic}

Let the approximate dynamics be given by
\begin{equation}\label{eq:dynamics_approx}
    s^{k+1} = \sdyn(s^k, a_A^k, a_H^k)
    \enspace,
\end{equation}
where $s^t\in\mathcal{S}\subset\mathbb{R}^{\tilde{n}}$ and $a_i^t\in\mathcal{A}_i\subset\mathbb{R}^{\tilde{m}_i}$ are the state and  action in the simplified dynamics $\sdyn$.
The index $k$ is associated to a discrete time step that may be equal to the low-level time step or possibly different (typically coarser).
We generically assume that there exists a function ${g:\mathcal{X}\to\mathcal{S}}$ assigning a simplified state $s\in\mathcal{S}$ to every full state~$x\in\mathcal{X}\subset\mathbb{R}^{n}$.
The approximation is usually made seeking $\tilde{n} < n$ to improve tractability.
This can typically be achieved by ignoring dynamic modes in $f_i$ with comparatively small time constants.
For example, we may assume that vehicles can achieve any lateral velocity within a bounded range in one time step,
and treat it as an input instead of a state.

We model the dynamic game under feedback closed-loop information (both players' actions can depend on the current state $s$ but not on the state history), allowing the human driver to condition her choice of $a_H^k$ on the autonomous vehicle's current action $a_A^k$ at every time step $k$, resulting in a Stackelberg (or leader-follower) dynamic game \cite{Simaan1973a}.
We need not assume that the human is an ideal rational player, but can instead allow her action to be drawn from a probability distribution.
This is amenable to the use of human models learned through inverse optimal control methods \cite{Ziebart2008,waugh2010inverse},
and can also be used to naturally account for modeling inaccuracies, to the extent that the human driver's behavior will inevitably depart from the modeling assumptions \cite{fisac2018probabilistically}.

We generalize the well-defined feedback Stackelberg dynamic programming solution \cite{Basar1984} to the case in which one of the players, in this case the \emph{follower}, has a noisy decision rule: $p(a_H^k | s^k, a_A^k)$.
The autonomous vehicle, here in the role of the \emph{leader}, faces at each time step $k$ the nested optimization problem
of selecting the action with the highest Q value, which depends on the human's decision rule $p$, in turn affected by the human's own Q values: 
\begin{subequations}
\begin{align}\label{eq:DSG_DP}
    \max_{a_A^k}\; & 
     Q_A^k( s^k, a_A^k )\\
    \text{ s.t.}\;
    & p(a_H^k\mid s^k, a_A^k) =
    \pi_H\big[ Q_H^k(s^k, a_A^k, \cdot)\big](a_H^k)
\end{align}
\end{subequations}
where $Q_A^k$ and $Q_H^k$ are the state-action value functions at time step $k$,
and $\pi_H:L^\infty \to \Delta(\mathcal{A}_H)$ maps every utility function ${q:\mathcal{A}_H\to\mathbb{R}}$ to a probability distribution over $\mathcal{A}_H$.
A common example of $\pi_H$ (which we use in Section \ref{sec:results}) is a noisy rational Boltzmann policy, for which:
\begin{equation}
    P(a_H\mid s, a_A) \propto e^{Q_H(s, a_A, a_H)}
    \enspace.
\end{equation}

The values $Q_A^k$ and $Q_H^k$ are recursively obtained in backward time through successive application of the dynamic programming equations for $k = K, K-1,\hdots,0$:
\begin{subequations}
\begin{align}
    & \pi_A^*(s) := \arg\max_a Q_A^{k+1}(s,a)\enspace, \quad \forall s\in\mathcal{S}\\
    & a_H^i \sim \pi_H\big[ Q_H^i(s^i, a_A^i, \cdot)\big]
    \enspace, \quad i \in\{k,k+1\}\\
    &  Q_H^k(s^k,a_A^k,a_H^k) = \tilde r_H(s^k,a_A^k,a_H^k) + \notag \\&\qquad\qquad\qquad\mathbb{E}_{a_H^{k+1}}Q_H^{k+1}(s^{k+1},\pi_A^*(s^{k+1}),a_H^{k+1})\\
    & Q_A^k(s^k,a_A^k) = \mathbb{E}_{a_H^k} \tilde r_A(s^k,a_A^k,a_H^k)\! +\! Q_A^{k+1}(s^{k+1},\pi_A^*(s^{k+1}))
\end{align}
\end{subequations}
with $s^{k+1}$ from \eqref{eq:dynamics_approx} and letting $Q_A^{K+1} \equiv 0$, $Q_H^{K+1} \equiv 0$.

\begin{algorithm}[t] 
\caption{Feedback Stackelberg Dynamic Program\label{alg:DSG_DP}}
  \KwData{$\hat{r}_A(\hat s,\hat a_A,\hat a_H),\;
         \hat{r}_H(\hat s,\hat a_A,\hat a_H)$}
  \KwResult{$\hat{V}_A(\hat s,k)$, $\hat{V}_H(\hat s,k)$,
 		   $\hat a_A^*(\hat s,k)$, $\hat a_H^*(\hat s,k)$}
  \BlankLine
  Initialization\DontPrintSemicolon\;\PrintSemicolon
  \For{$\hat s \in \hat{\mathcal{S}}$}{
    \nlset{A0}
    $\hat{V}_A(\hat s, K+1) \gets 0$\;
    \nlset{H0}
    $\hat{V}_H(\hat s, K+1) \gets 0$\;
  }
  \BlankLine
  Backward recursion\DontPrintSemicolon\;\PrintSemicolon
  \For{${k}\gets K$ \KwTo $0$}{
    \For{$\hat s \in \hat{\mathcal{S}}$}{
      \For{$\hat a_A\in \hat{\mathcal{A}}_A$}{
      	\For{$\hat a_H\in  \hat{\mathcal{A}}_H$}{
          \nlset{H1}
          $q_H(\hat a_H) \gets
          \hat r_H(\hat s,\hat a_A,\hat a_H)
          $\DontPrintSemicolon\;\PrintSemicolon$\qquad\qquad
          + \;\hat{V}_H(\sdyn(\hat s,\hat a_A,
          	\hat a_H),k+1)$\;
        }
        \nlset{H2}
        $P(\hat a_H \mid \hat a_A) \gets
            \pi_H[q_H](\hat a_H)$\;
        \nlset{H3}
        $q_{H}^*(\hat a_A)\gets \sum_{\hat a_H} P(\hat a_H \mid \hat a_A) \times
            q_H(\hat a_H)$\;
        \nlset{A1}
        $q_A(\hat a_A) \gets
            \sum_{\hat a_H} P(\hat a_H \mid \hat a_A) \times
            $\DontPrintSemicolon\;\PrintSemicolon$\qquad\qquad
            \Big(
            \hat r_A(\hat s,\hat a_A,\hat a_H^*(\hat a_A))
            $\DontPrintSemicolon\;\PrintSemicolon$\qquad\qquad
            + \;\hat{V}_A(\sdyn(\hat s,\hat a_A,
        	  a_H^*(\hat a_A)),k+1)\Big) $\;
		  }
		  \nlset{A2}
      $\hat a_A^*(\hat s,k)\gets 
      \arg\max_{\hat a_A} q_A(\hat a_A)$\;
      \nlset{A3}
      $\hat{V}_A(\hat s,k)\gets
      q_A(\hat a_A^*(\hat s,k))$\;
      \nlset{H4}
      $\hat a_H^*(\hat s,k)\gets
      a_H^*(\hat a_A^*(\hat s,k))$\;
      \nlset{H5}
      $\hat{V}_H(\hat s,k)\gets
      q_H^*(\hat a_A^*(\hat s,k))$\;
    }
  }
\end{algorithm}

The solution approach is presented in Algorithm \ref{alg:DSG_DP} for a discretized state and action grid $\hat{\mathcal{S}}\times \hat{\mathcal{A}}_A \times \hat{\mathcal{A}}_H$.
This computation is typically intensive, with complexity $O\big(|\hat{\mathcal{S}}|\cdot|\hat{\mathcal{A}}_A|\cdot|\hat{\mathcal{A}}_H|\cdot K\big)$,
but is also extremely parallelizable,
since each grid element is independent of the rest and the entire grid can be updated simultaneously,
in theory permitting a time complexity of $O(K)$.
Although we precomputed the game-theoretic solution, our proposed computational method for the strategic planner can directly benefit from the ongoing advances in computer hardware for autonomous driving \cite{kenwell2018nvidia},
so we expect that it will be feasible to compute the strategic value in an online setting.

Once the solution to the game has been computed, rather than attempting to \emph{execute} any of the actions in this simplified dynamic representation, the autonomous vehicle can use the resulting value $V(s) := \max_a Q^0(s,a)$ as a guiding terminal reward term for the short-horizon trajectory planner.

\subsection{Tactical planner: Open-loop trajectory optimization}\label{subsec:tactical}

In this section we demonstrate how to incorporate the strategic value into a low-level trajectory planner.
We assume that the planner is performing a receding-horizon trajectory optimization scheme, as is commonly the case in state-of-the-art methods \cite{Pendleton2017}.
These methods tend to plan over relatively short time horizons (on the order of $1$ s), continually generating updated ``open-loop" plans from the current state---in most cases the optimization is local, and simplifying assumptions regarding the interaction are made in the interest of real-time computability.

While, arguably, strategic interactions can be expected to have a smaller effect over very short time-scales,
the vehicle's planning should be geared towards efficiency and safety beyond the reach of a single planning window.
The purpose of incorporating the computed strategic value is to guide the trajectory planner towards states from which desirable long-term performance can be achieved.

We therefore formalize the tactical trajectory planning problem as an optimization with an analogous objective to~\eqref{eq:return} with a shorter horizon $M<<N$ and instead introduce the strategic value as a terminal term representing an estimate of the optimal reward-to-go between $t=M$ and $t=N$:
\begin{equation}\label{eq:return_with_value}
    R_A(x^{0:M}, u_A^{0:M}, u_H^{0:M}) = \sum_{t=0}^{M} r_A(x^t, u_A^t, u_H^t) + V_A\big(g(x^t)\big)
    \enspace.
\end{equation}
The only modification with respect to a standard receding-horizon trajectory optimization scheme is the addition of the strategic value term.
Using the numerical grid computation presented in Section \ref{subsec:strategic}, this can be implemented as an efficient look-up table, allowing fast access to values and gradients (numerically approximated directly from the grid).

The low-level optimization of \eqref{eq:return_with_value} can thus be performed online by a trajectory optimization engine, based on some short-term predictive model of human decisions conditioned on the state and actions of the autonomous vehicle.
In our results we implement trajectory optimization similar to \cite{Sadigh2016} through a quasi-Newton scheme~\cite{andrew2007scalable},
in which the autonomous vehicle iteratively solves a nested optimization problem by estimating the human's best trajectory response to each candidate plan for the next $M$ steps.
We assume that the human has an analogous objective to the autonomous system, and can also estimate her strategic long-term value.
We stress, however, that our framework is more general,
and in essence agnostic to the concrete low-level trajectory optimizer used,
and other options are possible (e.g. \cite{fisac2018probabilistically,Schmerling2018}).

\section{Results}\label{sec:results}

In this section, we analyze the benefit of solving the dynamic game by comparing our hierarchical approach to using a tactical planner only, as in the state of the art \cite{Sadigh2016,Schmerling2018}. 
We then study what specific aspects of the hierarchical method lead to better performance, showcasing the importance of reasoning with the fully coupled information structure of the dynamic game.

\subsection{Implementation Details}

\subsubsection{Environment} We use a simulated two-lane highway environment with an autonomous car and human-driven vehicle. Both vehicles' rewards encode safety and efficiency (or progress) features, along with a preference for the left lane. For the purposes of these case studies, the autonomous car's reward also includes a target speed slightly faster than the human's and a preference for being ahead of the human.

\subsubsection{Tactical-Level Dynamics.} 
The dynamics of each vehicle are given by a dynamic bicycle model with a discrete time step $\Delta t=0.1$ s. The planner uses $M=5$ time steps.

\subsubsection{Strategic-Level Dynamics} 
The joint human-autonomous state space is 8-dimensional, making dynamic programming challenging. Our strategic level simplifies the state and dynamics using an approximate, high-level representation. 
We consider a larger time step of  \mbox{$\Delta k=0.5$ s} and a horizon \mbox{$K=10$} corresponding to \mbox{$5$ s}.
Since the environment is a straight highway, we  consider only the longitudinal position of the two vehicles relative to each other: $x_{\text{rel}} = x_\A - x_\H$. 
We additionally assume the human-driven vehicle's average velocity is close to the nominal highway speed \mbox{30~m/s}, and the vehicles' headings are approximately aligned with the road at all times.
Finally, given the large longitudinal velocity compared to any expected lateral velocity, we assume that vehicles can achieve any desired lateral velocity (constrained to a maximum absolute value of \mbox{$2.5$ m/s$^2$}) within one time step (which is consistent with a typical \mbox{$1.5$} s lane change).

To assess the effects of different degrees of simplification in the dynamics, we consider two different high-level models:

\noindent\textbf{\emph{Constant human lateral position}}. Assuming the human remains in the left lane, the high-level state becomes $[x_{\text{rel}}, y_A,v_{\text{rel}}]$ and the approximate dynamics are given by
\vspace{-0.3cm}
\begin{equation}\label{eq:dynamics_strategic}
    [\dot{x}_\text{rel},\: \dot{y}_A, \: \dot{v}_\text{rel} ] =
    [v_\text{rel},\: w_A,\: a_A - a_H - \tilde\alpha\cdot v_\text{rel} ]
    \enspace,
\end{equation}
with the control inputs being the autonomous car's lateral velocity $w_A$ and the vehicles' accelerations $a_A$, $a_H$,
and where $\tilde\alpha$ is the friction parameter.
This allows us to implement Algorithm \ref{alg:DSG_DP} on a \mbox{$101 \times 17 \times 43$} grid and compute the feedback Stackelberg solution of the high-level strategic~game.

\noindent\textbf{\emph{Dynamic human lateral position.}}
Removing the assumption that the human remains in the left lane, the augmented state becomes $[x_\text{rel},y_A, y_H,v_{\text{rel}}]$ and the approximate dynamics are 
\vspace{-0.3cm}
\begin{equation}\label{eq:dynamics_strategic_4D}
    [\dot{x}_\text{rel}, \: \dot{y}_A,\: \dot{y}_H,\: \dot{v}_\text{rel}] =
    [v_\text{rel},\: w_A,\: w_H,\: a_A - a_H - \tilde\alpha\cdot v_\text{rel}]
\enspace.
\end{equation}
We can then compute the feedback Stackelberg solution on a $75 \times 12 \times 12 \times 21$ grid using Algorithm \ref{alg:DSG_DP}.

\subsubsection{Human simulation.} 
For consistency across our case studies, we simulated the human driver's behavior.
We found that for the simple maneuvers considered,
a simple low-level trajectory optimizer model produced
sufficiently realistic driving behavior.
We assume that the human driver makes accurate predictions of the autonomous vehicle's imminent state trajectory for the next $0.5$~s.

\subsection{Main Case Studies}

The main case studies compare the tactical short-horizon, hierarchical with constant human lateral position, and hierarchical with dynamic human lateral position planning methods for 3 different driving scenarios. For each scenario, we define a successful maneuver as the autonomous car overtaking and merging in front of the human-driven car.

We choose not to include the local human response estimate proposed in \cite{Sadigh2016} into the autonomous vehicle's tactical plan computation, captured through the second term in the ``total derivative"
\begin{equation}
\label{eq:gradient}
\frac{dR_A}{d\mathbf{u}_A} = \frac{\partial R_A}{\partial \mathbf{u}_A} + \frac{\partial R_A}{\partial \mathbf{u}_H} \frac{\partial \mathbf{u}_H^*}{\partial \mathbf{u}_A}
\enspace .
\end{equation}
Instead, we assume, at each gradient step that $\frac{\partial \mathbf{u}_H^*}{\partial \mathbf{u}_A}\simeq 0$ and
allow the optimization to proceed by iterated local best response (or iterated gradient play) between candidate plans and predicted human trajectories.
This assumption makes the reward computationally easier to optimize and empirically causes the autonomous car to behave less aggressively. We address this observation in further detail in Section \ref{subsub:hessian_influence}.

\subsubsection{Easy Merge}
We first study a typical merge maneuver where the autonomous vehicle is \emph{ahead} of the human in the right lane. The tactical planner causes the autonomous car to slowly merge in front of the human, while both hierarchical planners result in the car quickly merging in front of the human, thus receiving the larger left lane reward sooner.

\subsubsection{Hard Merge}
\begin{figure}
    \centering
    \includegraphics[scale = 0.288]{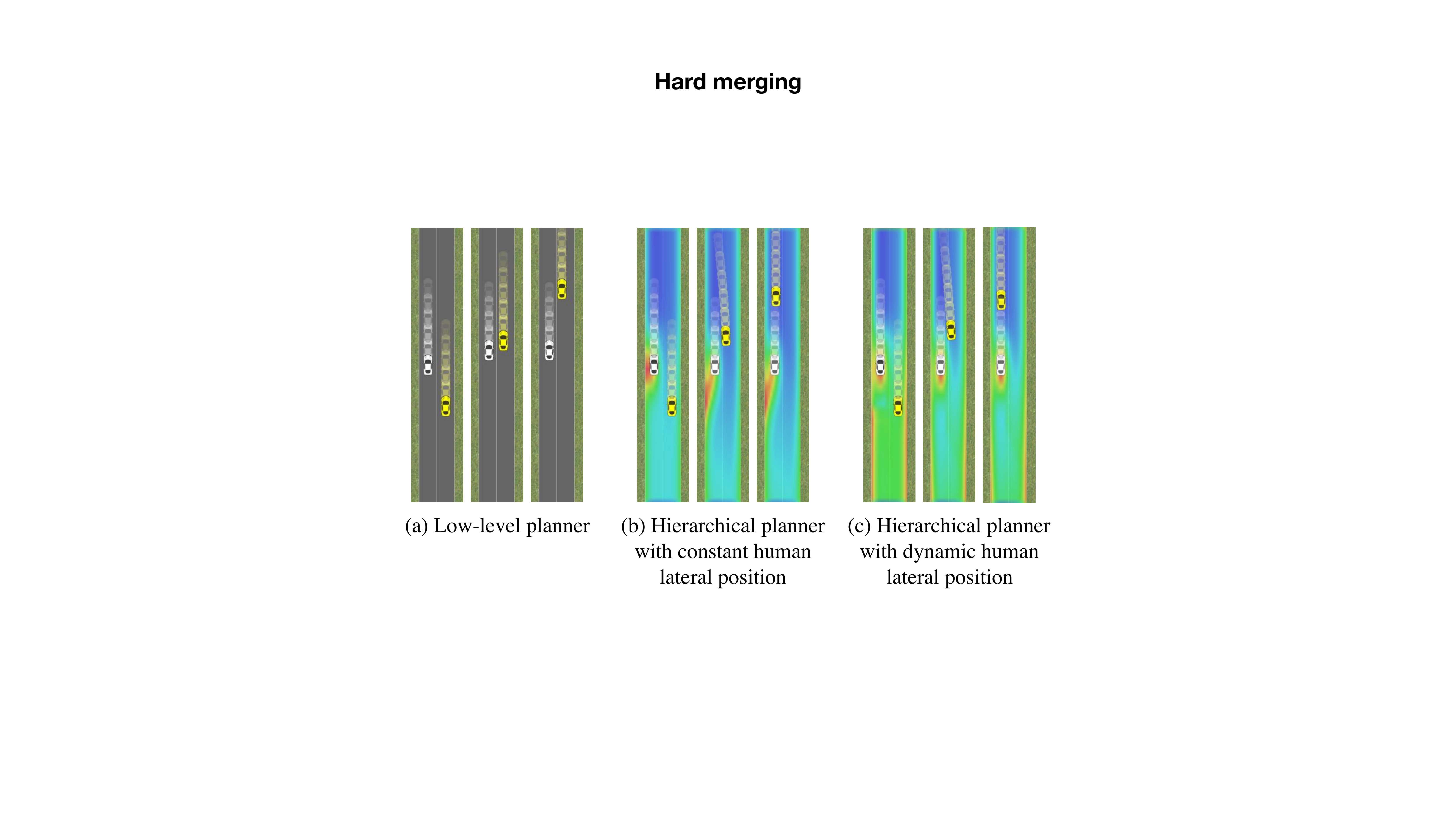}
    \caption{
    Comparison of the planners for the hard merging scenario. 
    The low-level trajectory planner (a) overtakes but does not merge into the left lane, while the game-theoretic hierarchical planners (b) and (c) successfully merge in front of the human.}
    \label{fig:hard_merging}
\end{figure}

Next, we study a more difficult merge maneuver in which the autonomous car starts \emph{behind} the human in the right lane, depicted in Fig.~\ref{fig:hard_merging}. The tactical autonomous car overtakes the human but does not merge into the left lane. In contrast, both hierarchical autonomous cars overtake and merge in front of the human.

\subsubsection{Overtaking}
\begin{figure}[h]
    \centering
    \includegraphics[scale = 0.29]{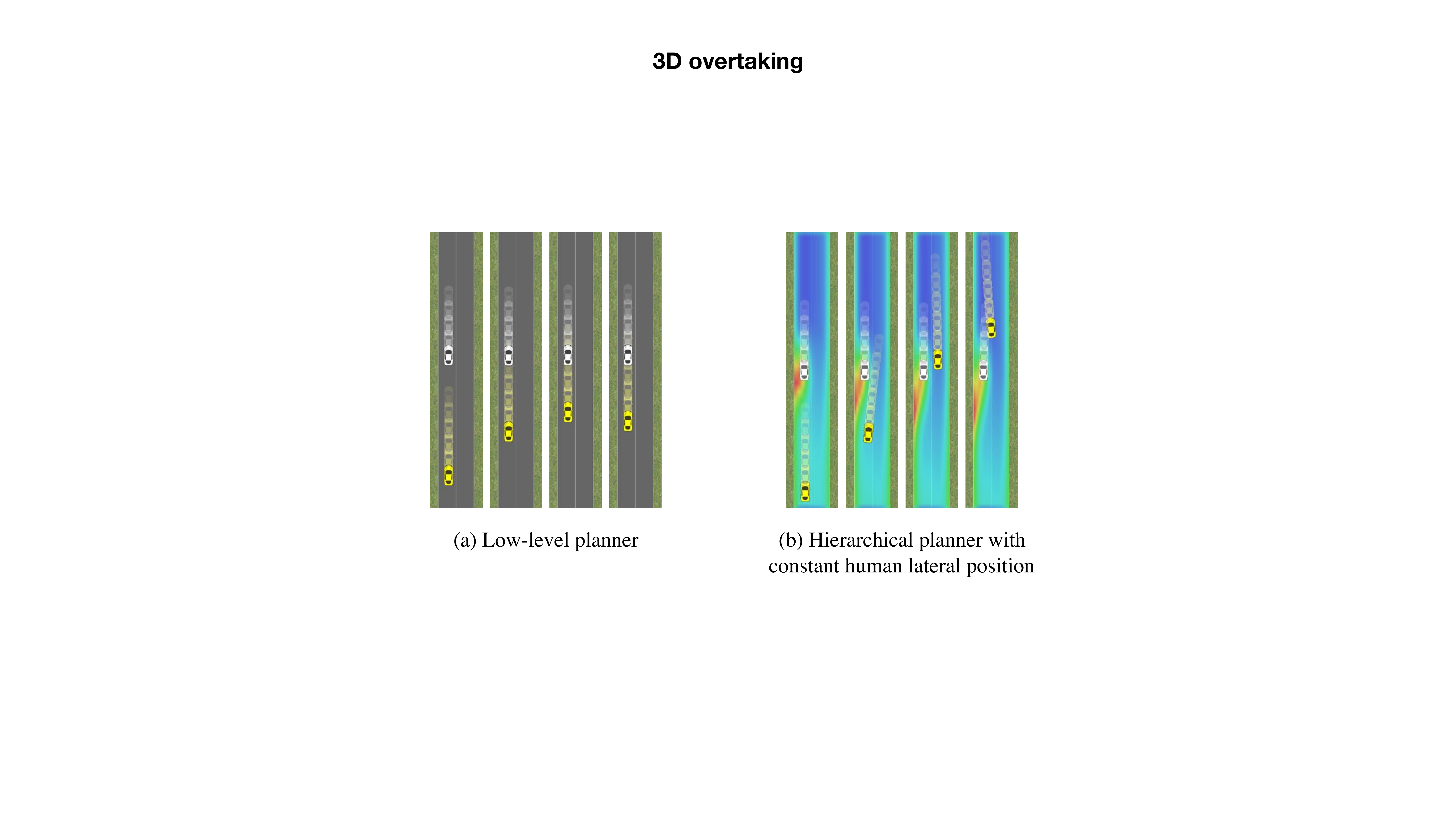}
    \caption{
    Demonstration of the tactical and hierarchical planners for the hard merging scenario. 
    The low-level trajectory planner (a) accelerates, but then brakes and remains behind the human.
    On the other hand, the increasing strategic value in the right lane incentivizes the hierarchical car (b) to change lanes, allowing it to overtake and merge in front of the human.}\vspace{-0.3cm}
    \label{fig:3d_overtaking}
\end{figure}

Finally, we study a complete overtaking maneuver in which the autonomous car starts \emph{behind} the human in the same lane. 
The tactical autonomous car does not successfully complete the maneuver---it first accelerates but then brakes to remain behind the human.
However, the hierarchical autonomous car with constant human lateral position successfully merges into the right lane, accelerates to overtake the human (reaching a maximum speed of 39.94 m/s, which is 4.94 m/s above its target speed), and merges back into the left lane, as shown in Fig.~\ref{fig:3d_overtaking}.
The hierarchical planner with dynamic human lateral position uses a more expressive high-level abstraction, which allows it to consider two alternative strategies, which are shown in Fig.~\ref{fig:4d_overtaking}.
The autonomous vehicle first approaches the human from behind, expecting her to have an incentive (based on her strategic value) to change lanes and let it pass.
If this initial ``tailgating'' strategy is successful and the human changes lane, the autonomous vehicle overtakes from the left lane.
Conversely, if the human does not begin a lane change, the autonomous vehicle is in turn incentivized to execute the same
``lane-change'' strategy exhibited by the hierarchical planner with constant human lateral position.

\subsection{In-depth Analysis}
We observe better results from hierarchical game-theoretic planning; we now seek to investigate \emph{why}. Is the strategic value merely lengthening the implicit horizon, helping escape local or myopic optima, or is the difference in information structure important? We find evidence for the latter below.

\subsubsection{Hierarchical vs. long-horizon and more global tactical planning}
\label{subsub:long-horizon}
The hierarchical planning method provides the autonomous car with more information about the future via the strategic value of the long-term game, which guides the optimization to escape local optima. If those were the only benefits, extending the horizon of the tactical planner and re-initializing in different basins of attraction ought to perform similarly. We thus extend the horizon to 2 s (20 time steps) and perform multiple independent optimizations initialized from diverse trajectories for each car: full-left steer, full-right steer, and straight steer (with acceleration input to maintain speed). This stronger tactical planner is unable to optimize in real time, unlike our other demonstrations, but is a good tool for analysis; extension beyond $2$ s was not tractable.

\begin{figure}[b]
    \centering
    \includegraphics[scale = 0.29] {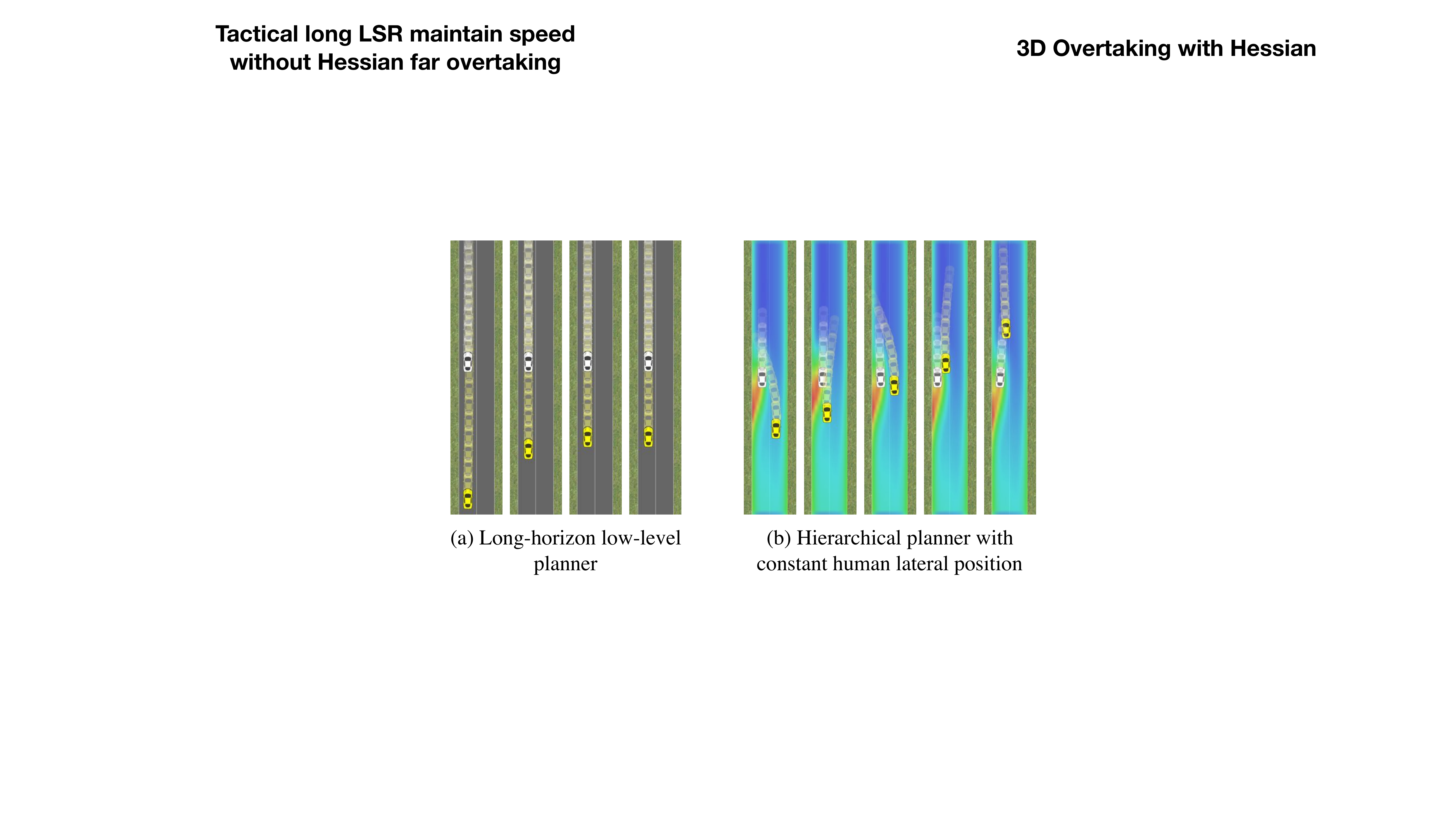}
    \caption{
    Demonstration of the tactical and hierarchical planners for the hard merging scenario. 
    (a) In the overtaking scenario, the long-horizon and more global tactical car accelerates, expecting the human to accommodate its higher speed to avoid a collision. 
    This forces the human to accelerate, and the autonomous car remains behind the human.
    (b) Using the influence term in the low-level trajectory optimization causes the hierarchical car to drive aggressively in the hard merging scenario, resulting in a collision.
    }
    \label{fig:long_horizon_low_level_and_hessian}
\end{figure}

We tested this planner in the overtaking scenario alongside a human-driven car that is aware of the autonomous car's plan, which is this planner's assumed information structure.
The planner still fails to complete the maneuver regardless of the initialization scheme and whether the influence term in \cite{Sadigh2016} is used, resulting in the autonomous car remaining behind the human, as shown in Fig.~\ref{fig:long_horizon_low_level_and_hessian}.
Moreover, we tested this planner against a human driver who maintains a constant slow speed of \mbox{$24$~m/s}.
In this case, the autonomous car brakes abruptly to avoid a collision and remains behind the human, at each time step expecting her to maximally accelerate for the next \mbox{$1$ s}.
Despite the longer horizon and more global optimization, this new tactical planner still assumes the wrong information structure, i.e. that the human knows the autonomous car's trajectory multiple seconds into the future.
This causes poor performance when the human does not in fact adapt to the autonomous vehicle's plan ahead of time.

\subsubsection{Differentiating through human response}
\label{subsub:hessian_influence}
When optimizing the autonomous car's trajectory at the tactical level, we computed a local equilibrium in trajectory space using iterated local best response.
We did not execute the implicit differentiation proposed in \cite{Sadigh2016}, by which the autonomous planner estimates the influence of each local trajectory change on the optimal response of the human.
We observed that incorporating this influence term resulted in more aggressive behavior in certain situations.
In the hard merge scenario, both hierarchical cars (with constant and dynamic human lateral position) attempted to merge into the left lane before having overtaken the human, resulting in a collision, as shown in Fig.~\ref{fig:long_horizon_low_level_and_hessian}.
In the overtaking scenario, the hierarchical car with constant human lateral position exhibited the  same aggressive behavior.

The above results seem to confirm that modeling the human driver as a ``pure" follower adapting to the autonomous vehicle's planned trajectory in an open-loop fashion, as formulated in \cite{Sadigh2016},
may lead to undesirably aggressive behavior.
Assuming that the human driver can accurately anticipate the autonomous car's planned trajectory can lead the autonomous vehicle to over-confidently execute actions that may lead to unsafe situations when the actual human driver fails to preemptively make way for it.

In other situations, including the influence term had no effect on the autonomous vehicle's behavior.
In view of the above, and given that computing this term is substantially more expensive,
we find that omitting it is a better choice for these driving scenarios.

\subsubsection{Confidence in Strategic Human Model}
In this section, we study the effects of varying the autonomous planner's confidence in its high-level model of the human.
By modeling the human as a Boltzmann noisily rational agent, we can naturally incorporate the autonomous planner's confidence in the human model via the inverse temperature parameter, as  in~\cite{fisac2018probabilistically}.
We can then compute different strategic values corresponding to varying levels of confidence in the human model.
In the overtaking scenario, we observed that using a strategic value with 1 as the inverse temperature parameter (as in the previous case studies) caused the autonomous car to successfully overtake the human, while inverse temperature values less than 0.3 result in the autonomous car remaining behind the human car.
A lower level of confidence in the human model discourages the autonomous car from overtaking because the human driver is more likely to act in an unpredictable manner that risks a collision.

\section{Discussion}\label{sec:conclusion}

We have introduced a hierarchical trajectory planning formulation for an autonomous vehicle interacting with a human-driven vehicle on the road.
To tractably reason about the mutual influence between the human and the autonomous system,
our framework uses a lower-order approximate dynamical model solve a nonzero sum game with closed-loop feedback information.
The value of this game is then used to inform the planning and predictions of the autonomous vehicle's low-level trajectory planner.

Even with a simplified dynamical model, solving the dynamic game will generally be computationally intensive.
We note, however, that our high-level computation presents two key favorable characteristics for online usability.
First, it is ``massively parallel" in the sense that all states on the discretized grid may be updated simultaneously.
The need for reliable real-time perception in autonomous driving has spurred the development of high-performance parallel computing hardware, which will directly benefit our method.
Second, once computed, the strategic value can be readily stored as a look-up table, enabling fast access by the low-level trajectory planner.
Of course, strategic values would need to be pre-computed for a number of scenarios that autonomous vehicles might encounter.

We believe that our new framework can work in conjunction with and significantly enhance existing autonomous driving planners, allowing autonomous vehicles to more safely and efficiently interact with human drivers.






\balance
\printbibliography

\end{document}